\documentclass[runningheads]{llncs}
\usepackage{graphicx}
\usepackage{authblk}
\usepackage{graphicx}
\usepackage{subfigure}
\usepackage{array}
\usepackage{float}
\usepackage{booktabs}
\usepackage{bbding}
\usepackage{etoolbox}
\usepackage{color}
\usepackage{soul}
\usepackage{algorithm}
\usepackage{algorithmic}
\usepackage{tabularx}
\usepackage{cite}
\usepackage{balance}
\usepackage{amsmath, amssymb}
\usepackage{makecell}
\usepackage{bm}
\usepackage{multirow}
\usepackage{color,xcolor}
\usepackage{hyperref}
\usepackage{colortbl}
\hypersetup{
    colorlinks,
    linkcolor={blue!50!black},
    citecolor={blue!50!black},
    urlcolor={blue!50!black}
}

\usepackage[T1]{fontenc}
\usepackage{verbatim}
\begin{document}
\title{Endo-4DGX: Robust Endoscopic Scene Reconstruction and Illumination Correction with Gaussian Splatting}
%
\titlerunning{Endo-4DGX}
\authorrunning{Y. Huang et al.}
\author{Yiming Huang\inst{1, 6~\star}
\and Long Bai\inst{1, 6~\star}
\and Beilei Cui\inst{1, 6}
\thanks{Co-first authors.}
\and Yanheng Li\inst{5}
\and Tong Chen\inst{4}
\and Jie Wang\inst{6}
\and Jinlin Wu\inst{5}
\and Zhen Lei\inst{5}
\and Hongbin Liu\inst{5}
\and Hongliang Ren\inst{1,2,6}
\thanks{Corresponding author.}}
\institute{Department of Electronic Engineering, The Chinese University of Hong Kong (CUHK), Hong Kong SAR, China
\and Shun Hing Institute of Advanced Engineering, CUHK, Hong Kong SAR, China \and City University of Hong Kong, Hong Kong, China \and The University of Sydney, Sydney, NSW, Australia \and Centre for Artificial Intelligence and Robotics (CAIR), Hong Kong Institute of Science \& Innovation, Chinese Academy of Sciences, Hong Kong, China
\and Shenzhen Research Institute, CUHK, Shenzhen, China\\
\email{\{yhuangdl, b.long, beileicui\}@link.cuhk.edu.hk, hren@cuhk.edu.hk}}



\maketitle              
\begin{abstract}

Accurate reconstruction of soft tissue is crucial for advancing automation in image-guided robotic surgery. The recent 3D Gaussian Splatting (3DGS) techniques and their variants, 4DGS, achieve high-quality renderings of dynamic surgical scenes in real-time. However, 3D-GS-based methods still struggle in scenarios with varying illumination, such as low light and over-exposure. Training 3D-GS in such extreme light conditions leads to severe optimization problems and devastating rendering quality. To address these challenges, we present Endo-4DGX, a novel reconstruction method with illumination-adaptive Gaussian Splatting designed specifically for endoscopic scenes with uneven lighting. By incorporating illumination embeddings, our method effectively models view-dependent brightness variations. We introduce a region-aware enhancement module to model the sub-area lightness at the Gaussian level and a spatial-aware adjustment module to learn the view-consistent brightness adjustment. With the illumination adaptive design, Endo-4DGX achieves superior rendering performance under both low-light and over-exposure conditions while maintaining geometric accuracy. Additionally, we employ an exposure control loss to restore the appearance from adverse exposure to the normal level for illumination-adaptive optimization. Experimental results demonstrate that Endo-4DGX significantly outperforms combinations of state-of-the-art reconstruction and restoration methods in challenging lighting environments, underscoring its potential to advance robot-assisted surgical applications. Our code is available at~\url{https://github.com/lastbasket/Endo-4DGX}.

\keywords{ 3D Reconstruction  \and Illumination Correction \and  Robotic Surgery.}
\end{abstract}
\section{Introduction}

Endoscopic procedures have emerged as a cornerstone of medical practice, particularly in the context of minimally invasive surgeries. In the recent development of robot-assisted minimally invasive surgeries, accurate reconstruction of the endoscopic scene is of the utmost significance. The novel reconstruction and rendering techniques like NeRF~\cite{mildenhall2021nerf, zha2023endosurf, wang2022neural} and 3D Gaussian Splatting (3DGS)~\cite{kerbl20233d, yang2024deformable} achieves superior performance for real-time renderings, aligning well with the surgical need for deformable surgical scenes reconstruction~\cite{huang2024endo, liu2024endogaussian, yang2024deform3dgs, huang2025advancing}.

Despite the notable achievements of existing methods, they face a formidable challenge in varying illumination within the surgical scenes. The endoscopic workspace is dynamic and unpredictable; illumination changes quickly due to factors such as the hardware issue of the light source and reflections from surgical instruments and tissue. Low-light conditions~\cite{ma2020cycle, moghtaderi2024endoscopic} can make endoscope difficult to capture clear details of the soft tissue, while over-exposure~\cite{yang2023learning} leads to information loss. Previous works~\cite{bai2023llcaps, ma2020cycle, mou2023global, chen2024lightdiff} have made progress in addressing 2D-level endoscopic illumination challenges. Garcia et al.~\cite{garcia2023multi} introduced a structure-aware network on endoscopic exposure correction (EC). Recently, EndoUIC~\cite{bai2024endouic} was proposed as a novel adaptive diffusion model for endoscopic EC.

\begin{figure}[!t]
    \centering
    \includegraphics[width=\linewidth]{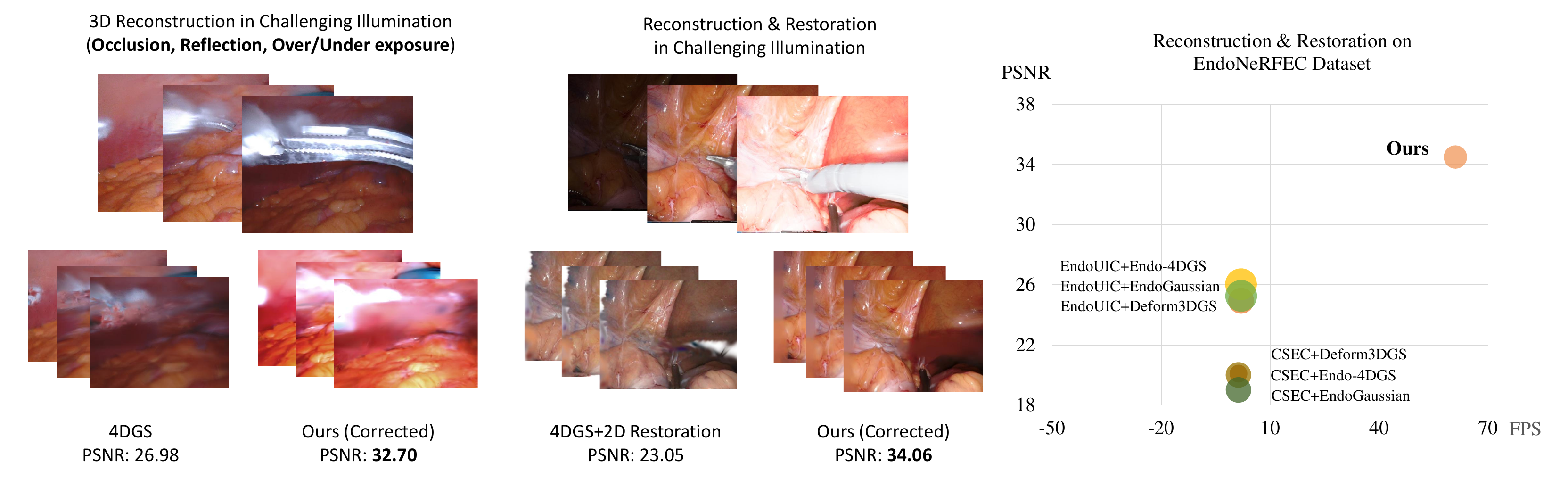}
    \caption{Our method achieves state-of-the-art performance for robust reconstruction and illumination correction in the surgical scene with varying illumination.}
    \label{fig:intro}
\end{figure}

However, due to the rapid variation of the surgical environment, 2D restoration methods~\cite{bai2023llcaps, garcia2023multi} fail to produce consistent results in real-time, leading to corrupted results for both appearance and geometry. While there exist 3DGS-based illumination correction solutions for general scenes~\cite{zhang2024darkgs, ye2024gaussian, kulhanek2024wildgaussians}, they are not applicable for surgical scenarios due to the lack of ability to model the sub-area and spatial-level illumination changes for deformable tissue, resulting in optimization failure of 3DGS. Besides, most general scene methods rely on pre-trained models~\cite{oquab2023dinov2} trained on daily life data~\cite{kulhanek2024wildgaussians,zuo2024fmgs}. In this case, using such a pre-trained model in surgical scenes shall disrupt the training of Gaussians due to the domain gap, leading to unsatisfactory outcomes when directly applying a generic scene approach to the surgical domain. Therefore, we propose Endo-4DGX, specifically designed for endoscopic reconstruction and illumination correction as shown in Fig.~\ref{fig:intro}. Endo-4DGX effectively models the view-dependent brightness variations by incorporating \textit{illumination embeddings}. To restore the reconstruction from challenging illumination, we develop two modules: The \textit{region-aware enhancement} module for Gaussian-level sub-area refinement. The \textit{spatial-aware adjustment} module for view-consistent exposure compensation across the entire image. Furthermore, we utilize \textit{global exposure control} loss to achieve illumination-adaptive optimization, enabling a more adaptable and practical reconstruction in real surgical settings. 
Specifically, our contributions are listed as follows:
\begin{itemize}
    \item We present Endo-4DGX, a novel endoscopic reconstruction method with illumination adaptive Gaussian Splatting. Our method achieves illumination correction and reconstruction in challenging uneven illumination.
    \item We introduce a region-aware enhancement module to resolve uneven lighting problems at the Gaussian level. Our region-aware module decodes the view-specific embedding to model the sub-area lightness changes.
    \item We design a spatial-aware adjustment module for spatial-level lightness adjustment, which focuses on the illumination refinement of the whole image.
    \item Our exhausted experiments on three real surgical datasets demonstrate the robustness of Endo-4DGX for surgical scene reconstruction under challenging illumination, providing a powerful tool for robot-assisted surgery.
\end{itemize}

\section{Methodology}
\begin{figure}[!t]
    \centering
    \includegraphics[width=\linewidth]{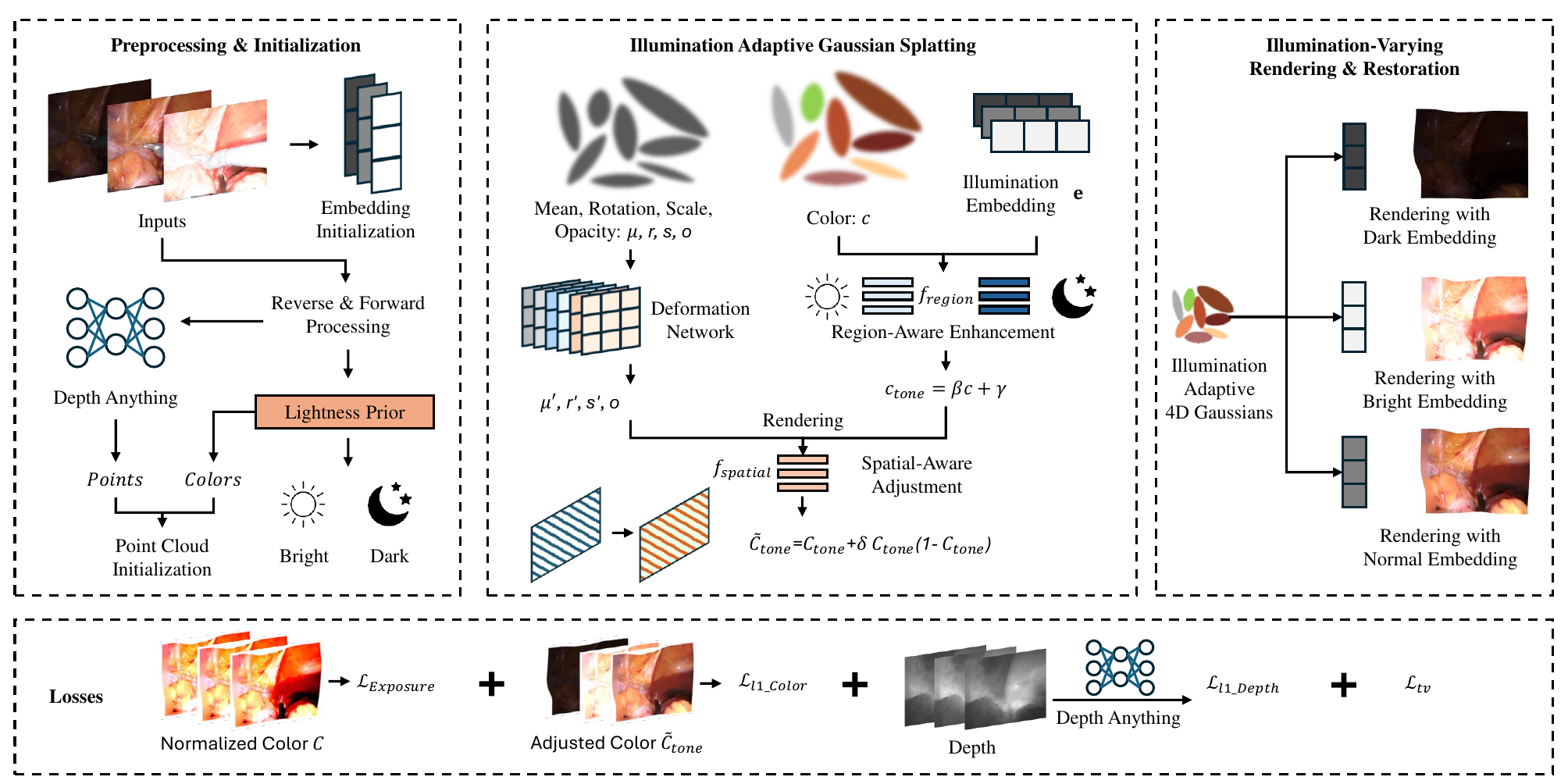}
    \caption{\textbf{Illustration of our proposed framework.} We first preprocess the inputs for initialization, and then we train the Gaussians with our illumination embedding, region-aware enhancement, and spatial-aware adjustment for sub-area and global restoration.}
    \label{fig:main}
\end{figure}

The overview of our proposed Endo-4DGX is shown in Fig~\ref{fig:main}. We first preprocess the input to obtain depth and illumination prior to initialization. Then, we train the Gaussians with our proposed \textit{region-aware enhancement} and \textit{spatial-aware adjustment} modules. Additionally, we normalize and optimize the rendering result with \textit{global exposure control}. We support the varying illumination rendering and the illumination correction with illumination embedding-based rendering.

\subsection{Preliminaries}\label{sec.2.1}

4D Gaussian Splatting (4DGS)~\cite{yang2024deformable, Wu_2024_CVPR} is an advanced technique for dynamic scene reconstruction, especially in endoscopic surgical applications~\cite{huang2024endo,liu2024endogaussian, yang2024deform3dgs}. We select 4DGS~\cite{Wu_2024_CVPR} as our baseline, which is built on 3D Gaussian Splatting~\cite{kerbl20233d} and represents a scene using 3D differentiable Gaussians for rapid rendering. Given a set of images and the corresponding camera poses, the deformable surgical scene can be represented by the 4DGS~\cite{Wu_2024_CVPR, huang2024endo, liu2024endogaussian} $\mathcal{G}^\prime=\{\mu^\prime, r^\prime, s^\prime, o^{\prime}, c\}$, where $\mu^\prime$, $r^\prime$, $s^\prime$,$o^\prime$ are the deformable mean, rotation, scale, and opacity, $c$ is the color. With the center depth $d_i$~\cite{huang2024endo}, the 2D covariance~\cite{zwicker2001surface} $\mathbf{\Sigma}^\prime$ and $\alpha = o^\prime\cdot e^{\frac{1}{2}(x-\mu^\prime)^T{\rm \mathbf{(\Sigma^\prime})}^{-1}(x-\mu^\prime)}$, the pixel-level color and depth is rendered by the alpha-blending~\cite{kerbl20233d} as:
\begin{equation}
    C(x) = \sum_i\alpha_i c_i \prod_{j=1}^{i-1} (1-\alpha_i), \ \ D(x) = \sum_i \alpha_i d_i \prod_{j=1}^{i-1} (1-\alpha_i),
\end{equation}


\subsection{Proposed Methodology: Endo-4DGX}\label{sec.2.2}
\noindent \textbf{Illumination Embedding and Region-Aware Enhancement.} We propose illumination embedding and region-aware enhancement to address the challenges posed by varying illumination conditions.  Before training, we pre-process all images $\{ I \}_N$ by a forward-reverse illumination estimation algorithm~\cite{zhang2019dual} $\mathcal{D}(\cdot)$. We utilize the pre-processed output $ \mathbf{p} =\mathcal{D}(I)$ as the lightness prior and classify the input images into two levels (Bright, Dark) for more precise and separated restoration. We define the illumination condition (IC) of each training image as:
\begin{equation}
    \text{IC}(I)=\left\{\begin{array}{ll}
    \text{Bright},&\text{if }\text{mean}(I)>\text{mean}(\mathbf{p})\\
    \text{Dark},&\text{if }\text{mean}(I)\leq \text{mean}(\mathbf{p})
    \end{array}\right.
\end{equation}
To deal with rapid illumination changes, we initialize trainable illumination embeddings $\mathbf{e}\in \mathbf{R}^{N\times k}$ for each specific time of view, in which $N$ is the number of training images, and $k$ is the embedding dimension which is set as 32 in our cases. During the training, all embeddings are updated with Gaussians jointly.

Inspired by~\cite{kulhanek2024wildgaussians}, we construct our region-aware enhancement as an affine transformation for Gaussian-level illumination enhancement. Given a view-specific color $C$, 
We input the view-specific illumination embedding into a lightweight concealing network $f_{region}$. However, using a single network for both the bright and dark image correction will lead to significant issues of training imbalance and convergence failure. To handle this, we proposed a unique approach involving two separate networks with different IC (Bright and Dark), where each network is formed by three layers of MLPs and sigmoid activations. By dividing the training images into separate ICs, we set the concealing network to bright or dark each time for enhancement. We split the output from the concealing network into two 1-channel enhancement parameters as $\beta, \gamma$. For a Gaussian's view-dependent color, the enhanced color is calculated as:
\begin{equation}
    c_{tone} = \beta \cdot c + \gamma, \ |  \ {\beta, \gamma} = f_{region}(c, \mathbf{e})
\end{equation}
Our illumination enhancement module models and fine-tunes the sub-area lightness of Gaussians based on the learned illumination embeddings and view-specific characteristics, achieving more accurate and realistic surgical scene reconstructions in the presence of varying illumination.
\\
\\
\noindent \textbf{Spatial-Aware Illumination Adjustment.}
While region-aware enhancement is useful for addressing illumination irregularities of individual elements or small regions within the scene, it lacks spatial-level adjustment ability, which makes it insufficient to tackle the broader challenges of image-level spatial illumination. To complement this, we further propose a spatial-aware illumination adjustment method. Inspired by~\cite{Zero-DCE}, we perform spatial adjustment by gathering the rendering output $C_{tone}$ from the regional enhanced color and generate a quadratic curve map with our spatial adjustment network $f_{spatial}$:
\begin{equation}
    \tilde{C}_{tone}=C_{tone} + \delta \cdot C_{tone}(1-C_{tone}) \ | \ \delta = f_{spatial}(\mathbf{e}),
\end{equation}
where $f_{spatial}$ is a lightweight network with three layers of MLPs and tanh activation.  By using this adjustment, we effectively fine-tune the illumination at the spatial level, making the overall scene more visually consistent.
\\

\noindent \textbf{Illumination-Adaptive Optimization.}
We introduce an illumination-adaptive optimization strategy to maintain stable training under varying illumination conditions. We utilize the L1 loss $\mathcal{L}_{l1}$ and Total Variation (TV) loss $\mathcal{L}_{tv}$ for the rendered color $\tilde{C}_{tone}$ and depth $D$ with illumination embedding. Following~\cite{huang2024endo}, we supervise the render depth with the ground truth depth $\hat{D}$ from stereo matching or pre-trained model~\cite{yang2024depth}. The loss of color and depth is defined as:
\begin{equation}
    \mathcal{L}_{color} = |\tilde{C}_{tone}-\hat{C}|\odot M + \lambda_{tv}L_{tv}(\tilde{C}_{tone}), 
\end{equation}
\begin{equation}
    \mathcal{L}_{depth} = \lambda_{depth}|\frac{D}{D_{\max}}-\frac{\hat{D}}{\hat{D}_{\max}}|\odot M+\lambda_{tv} L_{tv}(D),
\end{equation}
where $\lambda_{depth}, \lambda_{tv}$ are the weight for depth and tv loss, $M$ is the tool mask, and $\hat{C}$ is the ground truth color. Inspired by~\cite{Zero-DCE,cui_aleth_nerf}, we also propose a global exposure control loss for the rendered color $C$ without the illumination embedding:
\begin{equation}
    \mathcal{L}_{Exposure} = ||\text{avgpool1d}(\frac{1}{3}\sum_{j\in {R,G,B}}{C_j})-\mathbf{E}||^2,
\end{equation}
where $\mathbf{E}$ is the exposure level (0.6 is adopted), $\text{avgpool1d}$ is a 1-dimensional pooling. The global exposure control loss ensures that the reconstructed scene has a consistent and normal exposure despite the inputs being over- or under-exposure. The final loss is represented as:
\begin{equation}
    \mathcal{L} = \mathcal{L}_{color} + \mathcal{L}_{depth} + \mathcal{L}_{Exposure}
\end{equation}

\section{Experiments}

\begin{table}[!t]
\caption{
    \textbf{Quantitative results on EndoNeRF-EC Dataset~\cite{wang2022neural} for illumination correction}.    
  The \colorbox{red!40}{first}, \colorbox{orange!50}{second}, and \colorbox{yellow!50}{third} values are highlighted. * indicates that the method fails with NaN loss during reconstruction. Our method shows overall superior performance over state-of-the-art baseline methods.
    }
\centering%
\resizebox{0.95\textwidth}{!}{%
\begin{tabular}{l|c|c|c|c|c|c|c|c}
\toprule
        \multirow{2}{*}{Method}& \multirow{2}{*}{\shortstack{FPS}$\uparrow$}& \multirow{2}{*}{\shortstack{GPU}$\downarrow$} & \multicolumn{3}{c|}{Pulling} & \multicolumn{3}{c}{Cutting}  \\
         & & & PSNR$\uparrow$ & SSIM$\uparrow$ & LPIPS$\downarrow$ & PSNR$\uparrow$ & SSIM$\uparrow$ & LPIPS$\downarrow$  \\
        \midrule
    EndoUIC\cite{bai2024endouic}+Deform3DGS\cite{yang2024deform3dgs} & \textit{2.02} &8 GB&{27.15}&{0.892}&{0.180}&{22.74}&{0.842}&{0.231}\\
    EndoUIC\cite{bai2024endouic}+Endo-4DGS\cite{huang2024endo} & \textit{2.0}& 12 GB &\colorbox{orange!50}{28.79}& \colorbox{orange!50}{0.929} & \colorbox{orange!50}{0.097} & \colorbox{orange!50}{23.30} & \colorbox{orange!50}{0.890} & \colorbox{orange!50}{0.116}  \\
    EndoUIC\cite{bai2024endouic}+EndoGaussian\cite{liu2024endogaussian} & \textit{2.03} & 12 GB&\colorbox{yellow!50}{27.49}& \colorbox{yellow!50}{0.904} & 0.244 & \colorbox{yellow!50}{23.05} & \colorbox{yellow!50}{0.867} & 0.257  \\
    CSEC\cite{li2024color}+Deform3DGS\cite{yang2024deform3dgs} & \textit{1.29} & \colorbox{orange!50}{3.8 GB}&{20.03}& 0.855 & 0.194 & 20.18 & 0.809 & 0.251  \\
    CSEC\cite{li2024color}+Endo-4DGS\cite{huang2024endo} & \textit{1.27} & 7.8 GB&{19.89}& 0.876 & \colorbox{yellow!50}{0.140} & 20.14 & 0.838 & \colorbox{yellow!50}{0.186}  \\
    CSEC\cite{li2024color}+EndoGaussian\cite{huang2024endo} & \textit{1.29} & 7.8 GB&{18.63}& 0.838 & 0.283 & 19.41 & 0.800 & 0.339  \\
    DarkGS*\cite{zhang2024darkgs} & \colorbox{red!40}{524} & \colorbox{red!40}{2 GB}&6.04& 0.001 & 0.655 & 6.66 & 0.002 & 0.635  \\
    Gaussian-DK*\cite{ye2024gaussian} & \textit{41.0} &\colorbox{red!40}{2 GB} & 13.13 & 0.705 & 0.528 & 10.26 & 0.543 & 0.579  \\
    WildGaussians\cite{kulhanek2024wildgaussians} & \colorbox{orange!50}{108} & \colorbox{red!40}{2 GB} & 15.28 & 0.724 & 0.431 & 17.71 & 0.752 & 0.376  \\
    \textbf{Ours} & \colorbox{yellow!50}{61} & \colorbox{yellow!50}{6.5 GB}     &\colorbox{red!40}{34.94}&\colorbox{red!40}{0.946}&\colorbox{red!40}{0.048}&\colorbox{red!40}{34.06}&\colorbox{red!40}{0.950}&\colorbox{red!40}{0.043}\\
\bottomrule
    \end{tabular}
}


\label{tab:tab1}
    
\end{table}
\begin{table}[!t]
\centering%
\caption{
    \textbf{Quantitative results on StereoMIS~\cite{hayoz2023robust} and C3VD~\cite{bobrow2023} datasets for uneven illumination scene reconstruction}.
  The \colorbox{red!40}{first} and \colorbox{orange!50}{second} are highlighted. Our method surpasses all state-of-the-art methods.
    }
\resizebox{\textwidth}{!}{%
\begin{tabular}{l|c|c|c|c|c|c|c|c|c|c|c|c}
\toprule
\multirow{3}{*}{\shortstack{Method}}& \multicolumn{4}{c|}{EndoNeRF-EC} & \multicolumn{4}{c|}{StereoMIS} & \multicolumn{4}{c}{C3VD} \\
        & \multicolumn{2}{c|}{Pullling} & \multicolumn{2}{c|}{Cutting} & \multicolumn{2}{c|}{P1$\_$1} & \multicolumn{2}{c|}{P1$\_$2} & \multicolumn{2}{c|}{Cecum\_t2\_b} & \multicolumn{2}{c}{Sigmoid\_t2\_a} \\
        & PSNR$\uparrow$ & SSIM$\uparrow$ & PSNR$\uparrow$ & SSIM$\uparrow$ & PSNR$\uparrow$ & SSIM$\uparrow$ & PSNR$\uparrow$ & SSIM$\uparrow$ & PSNR$\uparrow$ & SSIM$\uparrow$ & PSNR$\uparrow$ & SSIM$\uparrow$ \\
        \midrule
    Deform3DGS\cite{yang2024deform3dgs} &\colorbox{orange!50}{30.63}&\colorbox{orange!50}{0.922}&{26.86}&{0.831}&\colorbox{orange!50}{28.55}&\colorbox{orange!50}{0.846}&{32.28}&{0.887}&\colorbox{orange!50}{33.07}&\colorbox{orange!50}{0.898}&{27.92}&{0.749}\\

    Endo-4DGS\cite{huang2024endo} &{28.51}& 0.909 & \colorbox{orange!50}{26.88} & \colorbox{orange!50}{0.850} & 26.98 & 0.817 & \colorbox{orange!50}{32.48} & 0.890 & 30.55 & 0.878 & \colorbox{orange!50}{32.03} & \colorbox{orange!50}{0.830} \\
    
    EndoGaussian\cite{liu2024endogaussian}  &{12.93}& 0.327 & 12.49 & 0.289 & 27.61 & 0.841 & 32.20 & \colorbox{orange!50}{0.896} & 5.10 & 0.001 & 5.99 & 0.001 \\
    \textbf{Ours}&\colorbox{red!40}{39.95}&\colorbox{red!40}{0.966}&\colorbox{red!40}{39.64}&\colorbox{red!40}{0.968}&\colorbox{red!40}{28.73}&\colorbox{red!40}{0.853}&\colorbox{red!40}{32.70}&\colorbox{red!40}{0.899}&\colorbox{red!40}{33.55}&\colorbox{red!40}{0.904}&\colorbox{red!40}{33.13}&\colorbox{red!40}{0.838}\\
\bottomrule
    \end{tabular}
}


\label{tab:tab2}
    
\end{table}
\subsection{Dataset}
\label{sec:dataset}

We split the training and testing sets as in~\cite{wang2022neural} and evaluate all baselines on an exposure correction (EC) dataset and two datasets with uneven illumination:\\
\textbf{EndoNeRF~\cite{wang2022neural} Exposure Correction (EndoNeRF-EC)} dataset includes two public video clips for robotic prostatectomy. The dataset is uniformly sampled as under-exposure, normal-exposure, and over-exposure in every 7 steps. We collaborate with professional photographers to simulate camera aperture adjustments using the Adobe Camera Raw SDK. The exposure values (EVs) are adjusted to analog the aperture changes. RAW images are rendered by different EV levels of three levels: under-exposure, over-exposure, and normal. Within each, the values are randomized, simulating real-world exposure mistakes.\\
\textbf{StereoMIS~\cite{hayoz2023robust}} dataset includes surgical video sequences for in-vivo porcine subjects. We select two video clips of frames from P1: 16400-16697 (P1\_1) and 7379-7499 (P1\_2), which contain the illumination variance due to the tool occlusion and reflection. We split the training and testing sets following~\cite{huang2024endo}. The ground truth depth is obtained by the stereo matching algorithm in OpenCV~\cite{opencv_library}.\\
\textbf{C3VD~\cite{bobrow2023}} dataset is a video data collection designed for cardiovascular reconstruction. We utilize the \textit{cecum\_t2\_b} and \textit{sigmoid\_t2\_a} clips, including uneven illumination caused by the light source direction and occlusions.

\subsection{Implementation Details}
\label{sec:implementation}

We evaluate the rendering quality by comparing the Peak Signal-to-Noise Ratio (PSNR), Structural Similarity Index (SSIM), and  Learned Perceptual Image Patch Similarity (LPIPS). We also report the Frame Per Second (FPS) and GPU memory usage to compare the computation. All methods are trained on a single RTX4090 GPU. We utilize the learning rate $1.6\times 10^{-3}$ following~\cite{liu2024endogaussian} with Adam optimizer. For the hyperparameters, we adopt $\lambda_{depth}, \lambda_{tv}=0.01$. To evaluate the illumination restoration, we utilize the first frame in the training data with normal-level illumination as the input embedding. While evaluating the uneven illumination reconstruction, we adopt the protocol in~\cite{kulhanek2024wildgaussians}. 
\begin{figure}[!t]
    \centering
    \includegraphics[width=\linewidth]{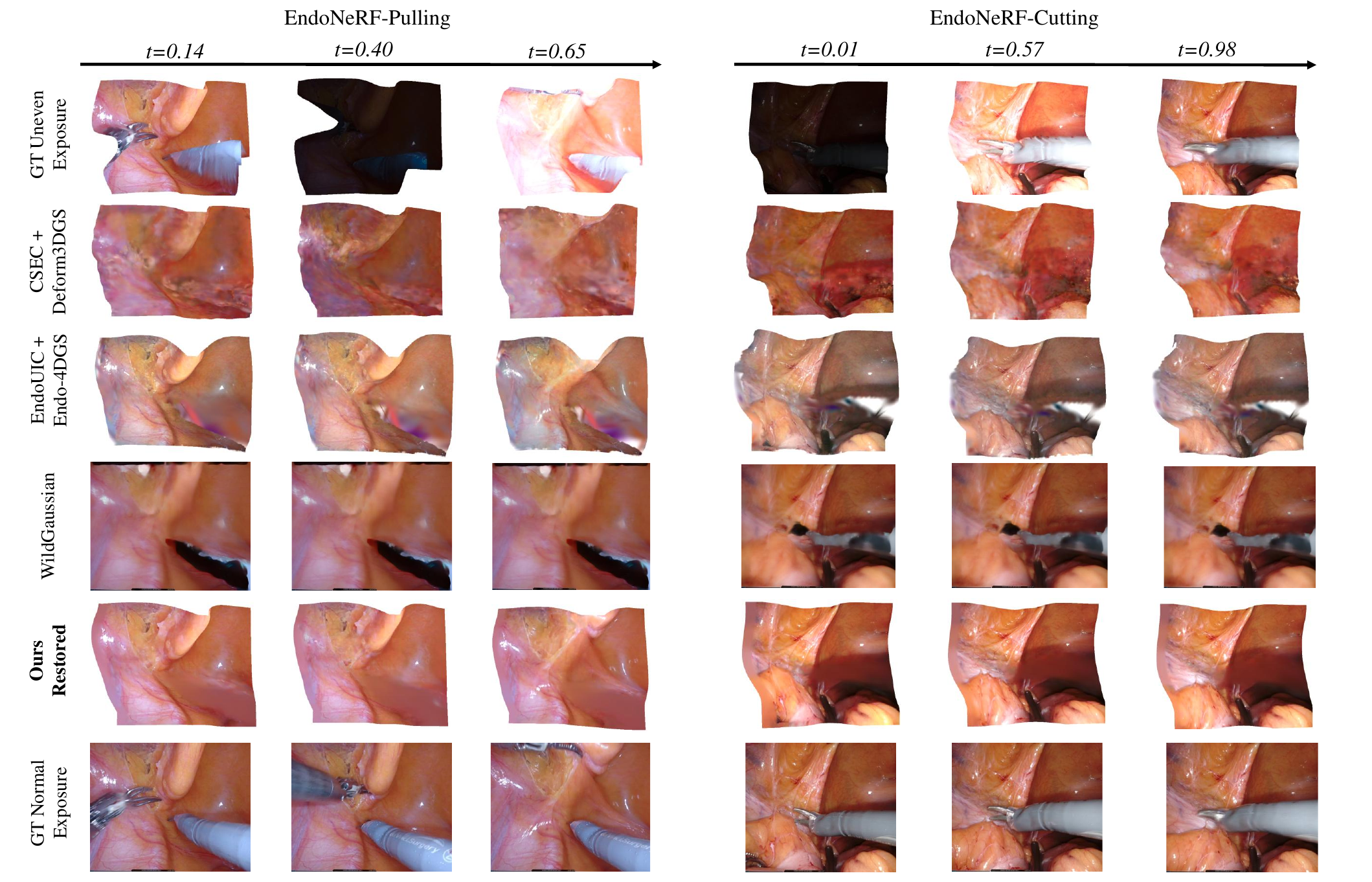}
    \caption{\textbf{Qualitative Result on EndoNeRF-EC~\cite{wang2022neural} Dataset.} Results for DarkGS and Gaussian-DK are not presented due to the reconstruction failure. Our method provides the best reconstruction and illumination correction results for challenging illumination.}
    \label{fig:qualitative}
\end{figure}
\subsection{Results}
We compare our method with three state-of-the-art Endoscopic Gaussian Splatting baselines~\cite{yang2024deform3dgs, huang2024endo, liu2024endogaussian} combined with state-of-the-art 2D illumination correction methods~\cite{bai2024endouic,li2024color}, and additionally three Gaussian Splatting-based methods for low-light enhancement~\cite{zhang2024darkgs, ye2024gaussian} and uneven illumination reconstruction~\cite{kulhanek2024wildgaussians}. For evaluation of~\cite{zhang2024darkgs}, we use the default configuration for da Vinci Xi system. Since there is no paired data for StereoMIS~\cite{hayoz2023robust} and C3VD~\cite{bobrow2023}, we evaluate the illumination correction on the EndoNeRF-EC~\cite{wang2022neural} dataset and evaluate the varying illumination reconstruction on all datasets. As demonstrated in Table~\ref{tab:tab1}, Endo-4DGX achieves state-of-the-art performance for the restoration task under challenging illumination conditions. Our method significantly outperforms existing reconstruction and restoration baselines for the illumination correction task on the EndoNeRF-EC dataset. Our method achieves PSNR/SSIM/LPIPS scores of 34.94/0.946/0.048 for the Pulling case and 34.06/0.950/0.043 for the Cutting case, surpassing the best baseline (EndoUIC~\cite{bai2024endouic}+Endo-4DGS~\cite{huang2024endo}) by over 6 dB in PSNR. Notably, while maintaining real-time rendering at 61 FPS, our approach reduces GPU memory consumption by 46\% compared to the combination of 2D restoration and 4DGS variants. In Table~\ref{tab:tab2}, despite illumination restoration, we conduct uneven illumination reconstruction on StereoMIS and C3VD datasets, where Endo-4DGX also demonstrates superior generalization with the highest PSNR. Additionally, our method achieves illumination correction under scenes without normal-level illumination embeddings by global exposure control. The qualitative result is shown in Fig~\ref{fig:qualitative}, demonstrating the improvement in terms of visual quality. For more visualization, please refer to the supplementary.

Ablation results in Table~\ref{tab:ablation} validate the necessity of each proposed component of our model. The results without illumination embedding drop significantly, even worse than the result without all modules, indicating that the integration of illumination embedding with region-aware enhancement and spatial-aware adjustment is necessary. Our full model outperforms the others without the proposed modules, confirming the critical role of each component in maintaining robust surgical scene reconstruction.
\begin{table}[t]
\centering
\caption{
    \textbf{Ablation experiments on EndoNeRF-EC dataset~\cite{wang2022neural}.} We evaluate the PSNR and SSIM of the varying illumination reconstruction and illumination correction by removing: (i) illumination embedding, (ii) region-aware enhancement, and (iii) spatial-aware adjustment. The \colorbox{red!40}{first} and \colorbox{orange!50}{second} result is highlighted.
}
\centering
\resizebox{\textwidth}{!}{
 \label{tab:ablation}
\begin{tabular}
{c|c|c|p{1.2cm}<{\centering} |p{1.2cm}<{\centering} | p{1.2cm}<{\centering} |p{1.2cm}<{\centering} |p{1.2cm}<{\centering} |p{1.2cm}<{\centering}| p{1.2cm}<{\centering} |p{1.2cm}<{\centering}}
\toprule

\multirow{2}{*}{\parbox{2cm}{\centering Illumination \\Embedding }} & \multirow{2}{*}{\parbox{2cm}{\centering Region-Aware\\Enhancement }} & \multirow{2}{*}{\parbox{2cm}{\centering Spatial-Aware\\Adjustment }} & 
\multicolumn{2}{c|}{Pulling} & \multicolumn{2}{c|}{Cutting} & \multicolumn{2}{c|}{Pulling-Correction} & \multicolumn{2}{c}{Cutting-Correction} \\ 

& & & PSNR $\uparrow$ & SSIM $\uparrow$  & PSNR $\uparrow$ & SSIM $\uparrow$ & PSNR $\uparrow$ & SSIM $\uparrow$ & PSNR $\uparrow$ & SSIM $\uparrow$\\ \midrule
\XSolidBrush      &\XSolidBrush      & \multicolumn{1}{c|}{\XSolidBrush}         & 26.95 & 0.892 &  27.59 & 0.856 & 22.88 & 0.770 &20.53 &0.734\\

\XSolidBrush       & \Checkmark     & \multicolumn{1}{c|}{\XSolidBrush}          & 10.40&0.535& 14.64 & 0.624 & 12.66 & 0.579 & 13.96 & 0.712\\

 \XSolidBrush      &\XSolidBrush      & \multicolumn{1}{c|}{\Checkmark}         & 24.56 & 0.840 & 30.53 & 0.895 & 23.46 & 0.815 & 19.93&0.692\\
 
\Checkmark       & \Checkmark     & \multicolumn{1}{c|}{\XSolidBrush}         & \colorbox{orange!50}{39.64} & \colorbox{orange!50}{0.965} & \colorbox{orange!50}{37.33} & \colorbox{orange!50}{0.957} & \colorbox{orange!50}{34.79} & \colorbox{orange!50}{0.945}& \colorbox{orange!50}{32.66}&\colorbox{orange!50}{0.933}\\

\Checkmark      &\XSolidBrush      & \multicolumn{1}{c|}{\Checkmark}         & 34.87 & 0.952 & 34.52  & 0.939 & 28.44 & 0.910& 25.09&0.865\\

\XSolidBrush       & \Checkmark    & \multicolumn{1}{c|}{\Checkmark}         & 14.36 & 0.627 & 31.06 & 0.925 & 21.94 & 0.804& 24.26&0.740\\

\Checkmark     & \Checkmark     & \multicolumn{1}{c|}{\Checkmark}         & \colorbox{red!40}{39.95} & \colorbox{red!40}{0.966} & \colorbox{red!40}{39.64} & \colorbox{red!40}{0.968} & \colorbox{red!40}{34.94} & \colorbox{red!40}{0.946} & \colorbox{red!40}{34.06} & \colorbox{red!40}{0.950}\\ \bottomrule
\end{tabular}}
\end{table}

\section{Conclusions}
We propose Endo-4DGX, an illumination-aware Gaussian Splatting framework for endoscopic 3D reconstruction under challenging lighting conditions. Our method achieves state-of-the-art performance for both reconstruction and illumination correction of dynamic lighting scenarios by unifying region-aware illumination enhancement and spatial-aware illumination adjustment. Experiments demonstrate our method outperforms existing methods while maintaining real-time rendering and robust generalization to various surgical datasets. This work advances practical applications in surgical navigation and robotic interventions, promoting the precision and safety of surgical procedures. Future research will extend the framework to broader clinical lighting challenges.

\begin{credits}
\subsubsection{Acknowledgements.}
This work was supported by Hong Kong RGC CRF C4026-21G,  RIF R4020-22, GRF 14211420, 14216020 \& 14203323).

\subsubsection{\discintname} The authors have no competing interests to declare that are
relevant to the content of this article.
\end{credits}
%
%
\bibliographystyle{splncs04}
\bibliography{reference}

\end{document}